\documentclass[conference]{IEEEtran}
\IEEEoverridecommandlockouts
\makeatletter
 \let\old@ps@headings\ps@headings
 \let\old@ps@IEEEtitlepagestyle\ps@IEEEtitlepagestyle
 \def\confheader#1{%

 \def\ps@IEEEtitlepagestyle{%
 \old@ps@IEEEtitlepagestyle%
 \def\@oddhead{\strut\hfill#1\hfill\strut}%
 \def\@evenhead{\strut\hfill#1\hfill\strut}%
 }%
 \ps@headings%
 }
 \makeatother

 \usepackage[pscoord]{eso-pic}

\usepackage{cite}
\usepackage{amsmath,amssymb,amsfonts}
\usepackage{algorithmic}
\usepackage{graphicx}
\usepackage{textcomp}
\usepackage{xcolor}

\makeatletter
\newcommand\subparagraph{\@startsection{subparagraph}{5}{\parindent}%
  {1ex \@plus1ex \@minus .2ex}%
  {-1em}%
  {\normalfont\normalsize}}
\makeatother
\usepackage{multirow}
\usepackage{lipsum}
\usepackage[utf8]{inputenc}
\usepackage{tikz}
\usetikzlibrary{shapes,arrows}
\usetikzlibrary{positioning}
\usepackage{subcaption}
\usepackage{tabularx}
\usepackage{subcaption}
\usepackage{fontspec}
\usepackage{polyglossia}
   \newfontfamily\bengalifont[Script=Bengali]{kalpurush.ttf}
\usepackage{caption}
\usepackage{enumitem}
\usepackage{array}
\usepackage[hidelinks]{hyperref}
\usepackage{eso-pic}

\DeclareCaptionLabelFormat{figlabel}{\textnormal{Fig. #2}}
\DeclareCaptionFormat{figformat}{\textnormal{#1#2#3}}
\def\BibTeX{{\rm B\kern-.05em{\sc i\kern-.025em b}\kern-.08em
    T\kern-.1667em\lower.7ex\hbox{E}\kern-.125emX}}
\begin{document}

\title{Open Vocabulary Word Recognition From Transcribed Bangla Texts \\
% {\footnotesize \textsuperscript{*}Note: Sub-titles are not captured in Xplore and
% should not be used}
% \thanks{978-1-6654-0906-3/21/\$31.00 ©2023 IEEE}
}

\author{\IEEEauthorblockN{Faias Satter}
\IEEEauthorblockA{\textit{Department of Computer Science and Engineering} \\
\textit{Khulna University of Engineering \& Technology }\\
 Khulna-9203, Bangladesh\\
faiaspromit@gmail.com}
\and
\IEEEauthorblockN{Sk. Md. Masudul Ahsan}
\IEEEauthorblockA{\textit{Department of Computer Science and Engineering} \\
\textit{Khulna University of Engineering \& Technology}\\
Khulna-9203, Bangladesh \\
smahsan@cse.kuet.ac.bd}

}
\author{\IEEEauthorblockN{Faias Satter\IEEEauthorrefmark{1}, Sk. Md. Masudul Ahsan\IEEEauthorrefmark{3}}
\IEEEauthorblockA{Department of Computer Science and Engineering\\
Khulna University of Engineering \& Technology\\
Khulna-9203, Bangladesh\\
Email: \IEEEauthorrefmark{1}faiaspromit@gmail.com \IEEEauthorrefmark{3}smahsan@cse.kuet.ac.bd}
}

\maketitle

% right after \begin{document}
\AddToShipoutPictureBG*{%
  \AtPageLowerLeft{%
    \raisebox{0.5cm}{\makebox[\paperwidth]{%
      \parbox[b]{0.85\paperwidth}{\scriptsize\centering
        \copyright~2023 IEEE. Personal use of this material is permitted. Permission from IEEE must be obtained for all other uses, in any current or future media, including reprinting/republishing this material for advertising or promotional purposes, creating new collective works, for resale or redistribution to servers or lists, or reuse of any copyrighted component of this work in other works.\\
        Accepted version of the paper published in the 2023 26th International Conference on Computer and Information Technology (ICCIT), DOI: 10.1109/ICCIT60459.2023.10441393.}}}}}

\begin{abstract}
An optical character recognition (OCR) can scan a paper and extract text using technology, making people's jobs easier. While various OCR systems are available in the software industry, finding a reliable equivalent solution for Bangla takes much work. When it comes to handwritten texts, the situation is much more unusual. Recognizing words from word images is the most critical stage in any OCR process. It is the second stage after segmenting words from text pictures. If this stage fails, the overall performance of the OCR will be poor, regardless of how well the other phases perform. This study aims to recognize words using deep learning in a handwritten Bangla word image. Three object detection models, SSD with MobileNetV2, Faster R-CNN with InceptionResNetV2, and an ensemble model of these two, have been used to train and test handwritten word images. A modified Non-Maximum Suppression has been introduced to enhance the effectiveness of the models' results. A customized dataset of 9841 handwritten Bangla word images has been compiled, featuring diverse handwriting styles from various individuals. All three models' performances have been checked against the test dataset, and the ensemble model has been the most impressive, with an F1-score of 92.61\%. Also, at the word level, the ensemble model correctly recognizes 96.12\% of the words to some extent. The system can be further improved by introducing a post-processing phase to correct errors generated by the system.
\end{abstract}


\section{Introduction}
One of the computer vision applications is optical character recognition (OCR), which translates images of typed/handwritten/printed text into machine-encoded text. OCR is commonly used to digitize texts in images to be electronically modified, searched, stored more compactly, and used in machine processes such as artificial intelligence, machine translation, text-to-speech, and document analysis and information extraction. However, there is currently no good OCR for the Bangla language. The most important stage in every OCR system is recognizing words from texts. It is the second stage after word segmentation from the texts. As a result, the authors intend to create a word recognition system specifically for recognizing handwritten Bangla words from photos of paper documents.

The authors aim to recognize words from an image of handwritten Bangla words by detecting the characters using three separate object detection models. The first one is Single Shot MultiBox Detector(SSD)\cite{id28}, a deep learning-based object detection model that recognizes characters using MobileNetV2\cite{id29} as a feature extractor. The second one is Faster R-CNN\cite{id26}, with InceptionResNetV2\cite{id27} as the feature extractor. The third one is an ensemble model, which considers both models' results and gives a prediction about the one with the highest confidence score.

After obtaining the models' predictions, a Non-Maximum Suppression algorithm is applied to refine the results for accuracy. The authors aim to propose modifications to the standard Non-Maximum Suppression algorithm to enhance its effectiveness. Subsequent to applying the modified Non-Maximum Suppression to the models' predictions, a process of rearrangement and character processing is undertaken to generate the final text representation of the words. However, handwritten Bangla characters are more complex than English characters. Many handwritten Bangla alphabets look so similar that it is difficult to distinguish them as unique characters. Moreover, characters in Bangla are connected by a line within a word called Matra, so they do not have separate spaces inside a handwritten Bangla word like in English. So, detecting different characters is a big challenge. Nevertheless, some characters can be written in multiple ways. This complication becomes more problematic if the work also includes conjunct characters. Also, different people have different handwriting styles and sizes. These limitations make developing a word recognition system for Bangla handwriting more complicated.

Most of the recent deep learning-based works done in this field have been on closed vocabulary, where the number of unique words has been fixed. The reason is the tendency to recognize the whole word as an individual entity. The authors intend to develop an open-vocabulary system with no limit to the number of unique words as it detects and recognizes characters from the word images and then tweaks and re-arranges the characters to give the word result in text. This process unseals the vocabulary limit, ensuring the system's effectiveness regardless of how many new/undiscovered words are introduced after training. Contrary to the limited works done so far on word recognition that typically focus on open vocabulary, where characters must be segmented from the word first, the authors have pursued a segmentation-free approach. This is particularly significant for Bangla handwriting, where a segmentation-based approach can be challenging due to its cursive nature.

To support this goal, the authors have created a vast dataset of handwritten Bangla word images where many words are non-words(e.g., words with combinations of characters that lack actual meaning and do not exist in the dictionary). This system aims to present the authors' efforts to develop a bespoke dataset and enhance word recognition performance from handwritten Bangla words. Although there is yet to be a high-quality OCR for Bangla like there is for English, Chinese, or Arabic, the authors hope their system will be a milestone toward meeting the demand for reliable OCR in Bangla. The scope of this paper encompasses the recognition of words in handwritten Bangla word images.

\section{Related Works}
Alom et al.\cite{id1} employed a Deep Convolutional Neural Network (DCNN) for Bangla character recognition. Their experiments showed that modern DCNN models like DenseNet, FractalNet, and ResNet outperformed other classical methods. Roy et al.\cite{id3} employed deep learning algorithms to recognize handwritten Bangla conjunct characters. They developed a novel DCNN using supervised layerwise training. They significantly reduced the error rate from 19\% to 9.67\%. Fardous et al.\cite{id4} developed a CNN-based model to distinguish handwritten solitary Bangla conjunct letters. They evaluated the model's performance by training it on CMATERdb 3.1.3.3. Purkayastha et al.\cite{id5} proposed a DCNN technique for recognizing Bangla handwritten characters. They tested their system on the BanglaLekha-Isolated dataset, achieving impressive accuracies of 89.93\% on practically all Bangla characters(80 character classes). However, the model's main challenge was distinguishing similar-form characters, resulting in most recognition mistakes. Ghosh et al.\cite{id6} presented a model for Bangla handwritten character recognition using the MobileNetV1 architecture. Their system recognized 231 classes, including 171 conjuncts, 50 basics, and 10 numerals, achieving an overall accuracy of 96.46\%. Adole et al.\cite{id10} investigated the application of Faster R-CNN InceptionResNetV2 on offline Kanji handwriting characters. They explored the effectiveness of this deep learning model for recognizing Kanji characters in handwritten form.

Islam et al.\cite{id7} introduced a Real-time Bangla license plate recognition system based on Faster R-CNN and SSD. The characters on the license plate are individually detected and then rearranged in their correct order. Their model achieved a precision of 91.67\% for character detection on the license plate. Abdullah et al.\cite{id8} developed a YOLO-based three-stage network for recognizing Bangla license plates in Dhaka. They used YOLOv3 to locate plates and ResNet-20-based CNN to recognize characters. Their method achieved over 92.7\% accuracy in character identification.

Sadaf et al.\cite{id32} introduced an end-to-end system for Bangla word recognition, leveraging pre-trained CNN architectures and recurrent units, including Puigcerver’s CRNN and Flor’s Gated-CNN. Their approach achieved superior performance with a CER of 12.83\% and a WER of 36.01\%. Dutta et al.\cite{id33} introduced an effective CNN-RNN hybrid architecture for lexicon-free handwritten text recognition in Devanagari and Bangla, surpassing prior methods on the RoyDB test set. Das et al.\cite{id34} introduced 'H-WordNet,' a holistic handwritten word recognition approach using a streamlined convolutional neural network. By sidestepping character-level segmentation and handcrafted features, the model achieved a significant parameter reduction while outperforming recent methods. Evaluation on a Bangla word database demonstrated H-WordNet's effectiveness with a recognition accuracy of 96.17\%.

Majid et al.\cite{id17} demonstrated a framework for ofﬂine handwriting recognition using character spotting and autonomous tagging, which works for any alphabetic script. Character spotting was built on the idea of object detection to ﬁnd character elements in unsegmented word images. In character spotting, they applied two Faster R-CNN with VGG-16 for the characters and the diacritics. Hasan et al.\cite{id18} used an object detection model, SSD with MobileNetV2, as a feature extractor to recognize characters from handwritten Bangla words. Their work did not include any conjunct character, and their approach correctly recognized only 39.67\% of the word images.

% related works jodi paro aro add koro

\section{Proposed Methodology}

% Flowchart
% Flowchart

\begin{figure}[b]
    \centering
    \resizebox{0.5\textwidth}{!}{
    \begin{tikzpicture}[node distance=0.4cm, every node/.style={rectangle, draw, minimum width=3cm, minimum height=0.75cm},
    double/.style={draw, anchor=text, rectangle split,rectangle split parts=2},
    triple/.style={draw, anchor=text, rectangle split,rectangle split parts=3},>=stealth']

      % Nodes
      \node (step1) {\textbf{1. Binary Word Images}};
      \node (step2) [below=of step1]{\textbf{2. Labeling Each Character}};
      \node (step3) [triple,below=of step2] {\textbf{3. Object Detection Models}
      \nodepart{second}
      SSD with MobileNetV2
      \nodepart{third}
      Faster R-CNN with InceptionResNetV2};
      \node (leftofstep3) [double,left=of step3] {
        \textbf{4. Train} 
        \nodepart{second}
        80\% words}; 
      \node (rightofstep3) [double,right=of step3] {
        \textbf{5. Result Acquisition From Models}  
        \nodepart{second}
        20\% words}; 
      \node (step5) [below=of rightofstep3] {\textbf{6. Applying Threshold on Class Confidence Scores}};
      \node (step6) [below=of step5] {\textbf{7. Non-Maximum Suppression(NMS) }};
      \node (step7) [below=of step6] {\textbf{8. Preserving Necessary Overlap(PNO)}};
      \node (step8) [below=of step7] {\textbf{9. Text Words }};
      % \node (step9) [below=of step8] {\textbf{9.  }};
      % Arrows
      \draw [->] (step1) -- (step2);
      \draw [->] (step2) -- (step3);
      \draw [->] (step3) -- (leftofstep3);
      \draw [->] (step3) -- (rightofstep3);
      \draw [->] (rightofstep3) -- (step5);
      \draw [->] (step5) -- (step6);
      \draw [->] (step6) -- (step7);
      \draw [->] (step7) -- (step8);
      % \draw [->] (step8) -- (step9);
    \end{tikzpicture}
    }
  \caption{Flowchart of the working procedure}
  \label{fig1}
\end{figure}

\subsection{Working Procedure}
The system's overall operation can be broken down into many steps. The working process is illustrated in Fig. \ref{fig1}. 
\subsubsection{Binary Word Images}
The handwritten Bangla words are taken input as binary images where character pixels are black and background pixels are white.(Fig. \ref{fig2a})

\subsubsection{Labeling Each Character}
\begin{figure}
  \centering
  \resizebox{0.3\textwidth}{!}{
      \begin{subfigure}[b]{0.49\textwidth}
        \centering
        \includegraphics[width=\textwidth,height=80pt]{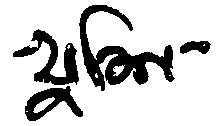}
        \captionsetup{font=LARGE}
        \caption{\LARGE Binary Word Image}
        \label{fig2a}
      \end{subfigure}
      \hfill
      \begin{subfigure}[b]{0.49\textwidth}
        \centering
        \includegraphics[width=\textwidth,height=80pt]{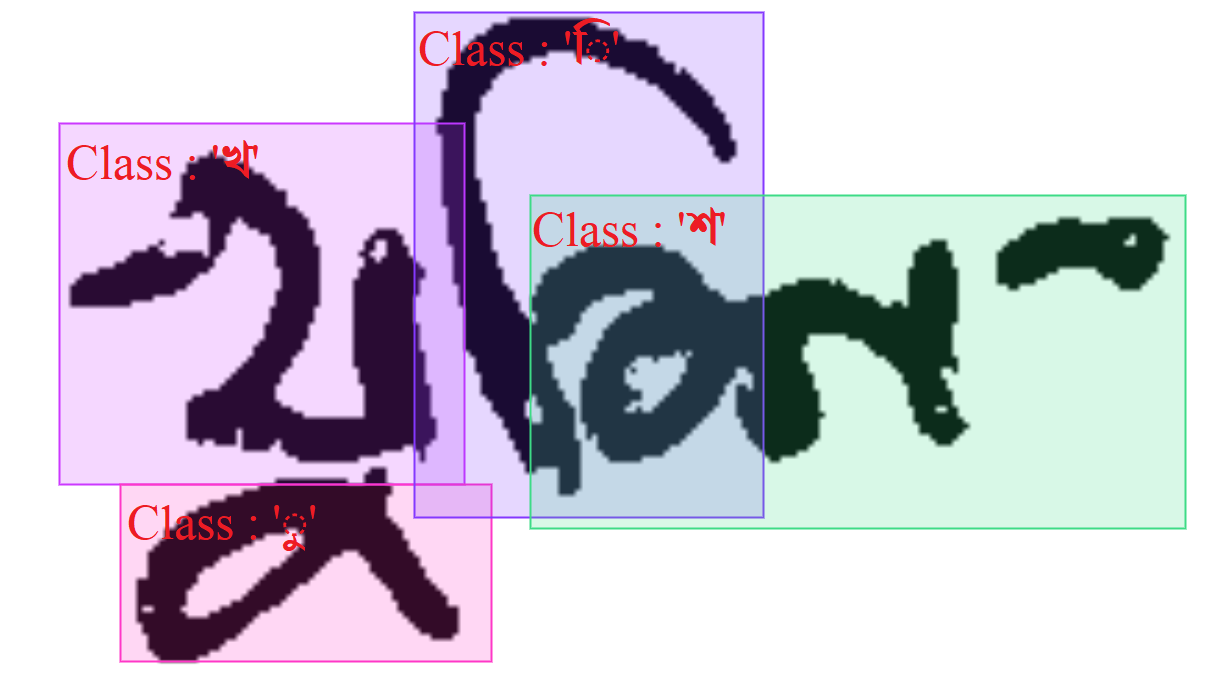}
        \captionsetup{font=LARGE}
        \caption{\LARGE Labeled Word Image}
        \label{fig2b}
      \end{subfigure}
      \hfill
    } %converted
  \caption{Binary image and character labels example}
  \label{fig2}
\end{figure}

All the characters/letters, including the vowels/diacritics from the word images, are labeled according to their respective classes. A total of 88 distinct classes(TABLE \ref{table1}), including the numbers, punctuation marks, and seventeen most used conjunct characters  \cite{id19}, have been used here for labeling. Fig. \ref{fig2b} shows how each character from the word images is labeled. Characters have been labeled after a rectangle bounding box has bounded them.

\subsubsection{Object Detection Models}
A total of 9841-word images have been labeled for this research, showcasing an open vocabulary. Among these images, nearly 80\% have been used to train the object detection models, while the remaining 20\% have been allocated for manipulating the models' results and evaluating the system's performance.

\begin{table}[b]
  \centering
  \caption{Different classes used in this research}
  \begin{tabularx}{0.5\textwidth}{|m{1.6cm}|m{6.56cm}|}
    \hline
    \centering Vowels & \centering \textbf{\bengalifont অ, ই, ঈ, উ, ঊ, ঋ, এ, ঐ, ও, ঔ} \tabularnewline
    \hline
    \centering Consonants & \centering \textbf{\bengalifont \textcolor{blue}{ক, খ, গ, ঘ,} ঙ, \textcolor{blue}{চ, ছ, জ, ঝ,} ঞ, \textcolor{blue}{ট, ঠ, ড, ঢ, ণ, ত, থ, দ, ধ, ন, প, ফ, ব, ভ, ম, য, র, ল, শ, ষ, স, হ, ড়, ঢ়, য়,} ৎ, ঃ, ং,\textcolor{blue}{ ঁ}}     \tabularnewline
    \hline
    \centering Numbers & \centering \textbf{\bengalifont ০, ১, ২, ৩, ৪, ৫, ৬, ৭, ৮, ৯}\tabularnewline
    \hline
    \centering Diacritics & \centering \textbf{\bengalifont া, ি, ী, ে, ু, ূ,ৈ, ৌ, ্র, ্য, ৃ}\tabularnewline
    \hline
    \centering Punctuations & \centering \textbf{\bengalifont ।}\tabularnewline
    \hline
    \centering Conjuncts & \centering \textbf{\bengalifont \textcolor{blue}{ক্ষ, ন্ত, ত্র, ঙ্গ, স্থ, স্ব, ক্ত, স্ত, ন্দ, চ্ছ, দ্ধ, ন্ত্র, ত্ত, ষ্ট, ন্ন, ল্প, ম্প}}\tabularnewline
    \hline
  \end{tabularx}
  
  \label{table1}
\end{table} 

\subsubsection{Train}
After gathering the training input data, the next step is to create the training configuration and label map. Two independent object-detection models, SSD with MobileNetV2 and Faster R-CNN with InceptionResNetV2, are employed in parallel for training. Tensorflow detection model zoo\cite{id20} is used to download the pre-trained models. Each model is fine-tuned using empirically determined optimum parameter values to accomplish its prescribed duties efficiently(TABLE \ref{table6}).

\begin{table}[t]
  \centering
  \caption{Training parameters applied on different models}
  \renewcommand{\arraystretch}{1.5}
  \begin{tabularx}{0.5\textwidth}{|m{2.80cm}|m{2.25cm}|m{2.70cm}|}
    \hline
    \centering Model & \centering SSD with MobileNetV2 & \centering Faster R-CNN with InceptionResNetV2 \tabularnewline
    \hline
    \centering Batch Size & \centering $16$ & \centering $1$  \tabularnewline
    \hline
    \centering No. of Classes & \centering $88$ & \centering $88$  \tabularnewline
    \hline
    \centering Initial Learning Rate & \centering $0.08$ & \centering $0.008$  \tabularnewline
    \hline
    \centering Optimizer & \centering Momentum Optimizer & \centering Momentum Optimizer  \tabularnewline
    \hline
    \centering Epochs & \centering $50000$ & \centering $50000$  \tabularnewline
    \hline
    
  \end{tabularx}
  
  \label{table6}
\end{table}

\subsubsection{Result Acquisition From Models}
\begin{figure}
  \centering
  \resizebox{0.5\textwidth}{!}{
      \begin{subfigure}[b]{0.45\textwidth}
        \centering
        \includegraphics[width=\textwidth,height=90pt]{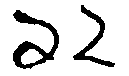}
        \captionsetup{font=Huge}
        \caption{\Huge Input}
        \label{fig5a}
      \end{subfigure}
      \begin{subfigure}[b]{0.45\textwidth}
        \centering
        \includegraphics[width=\textwidth,height=90pt]{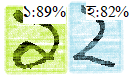}
        \captionsetup{font=Huge}
        \caption{\Huge SSD Result}
        \label{fig5b}
      \end{subfigure}
      \begin{subfigure}[b]{0.45\textwidth}
        \centering
        \includegraphics[width=\textwidth,height=90pt]{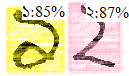}
        \captionsetup{font=Huge}
        \caption{\Huge Faster RCNN Result}
        \label{fig5c}
      \end{subfigure}
    % \resizebox{0.5\textwidth}{!}{
      \begin{subfigure}[b]{0.45\textwidth}
        \centering
        \includegraphics[width=\textwidth,height=90pt]{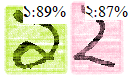}
        \captionsetup{font=Huge}
        \caption{\Huge Ensemble Result}
        \label{fig5d}
      \end{subfigure}
}
  \caption{Detected classes and confidence scores by SSD, Faster R-CNN, and Ensemble Model }
  \label{fig5}
\end{figure}

After conducting the models on images, characters are localized and recognized. The models generate rectangular bounding boxes, character classes, and corresponding class confidence scores. In this step, three different sets of results have been obtained. The first set of results is from the SSD model, the second is from the Faster R-CNN model, and the third is an ensemble result achieved through a multi-model ensemble approach. This ensemble result is obtained by merging the outputs of both the SSD and Faster R-CNN models and taking the best result for each character from the two. It results in a more comprehensive and accurate detection outcome. In Fig. \ref{fig5a}, for the word  \textbf{\bengalifont '১২'}, SSD detects \textbf{\bengalifont '১হ'}(Fig. \ref{fig5b}) where \textbf{\bengalifont '১'} has a confidence score of 89\% and \textbf{\bengalifont 'হ'} has 82\%. Similarly, Faster R-CNN detects the word as \textbf{\bengalifont '১২'} (Fig. \ref{fig5c}) with a confidence score of 85\% and 87\%. Ensemble model takes \textbf{\bengalifont '১'} from SSD and \textbf{\bengalifont '২'} from Faster R-CNN as they have the highest scores and gives  \textbf{\bengalifont '১২'}(Fig. \ref{fig5d}) as result. 

\subsubsection{Applying Threshold on Class Confidence Scores}
Each of the three models gives more than a hundred predictions for each character in the word. To refine these predictions, Non-Maximum Suppression(NMS) is applied first, reducing the number of predictions to a few for every character. Because, after training with 50,000 epochs, any class identified with a confidence score less than 20\% can be considered meaningless. For that, all classes with a confidence score of less than 20\%(empirically decided) are eliminated. Here in Fig. \ref{fig3b}, the characters' predicted classes, confidence scores, and bounding boxes by SSD are shown.

\subsubsection{Non-Maximum Suppression(NMS)}
Even after eliminating classes with a confidence score of less than 20\%, another round of NMS is applied to further refine the classes. This is necessary because, in the remaining results provided by the models, there may still be different predictions for the same character. For example, in Fig. \ref{fig3c}, SSD predicts three different classes \textbf{\bengalifont 'ষ,' 'ট,' 'ষ্ট'} for the same character \textbf{\bengalifont 'ষ্ট'}. A new term Overlap Value(OV) has been introduced to detect overlap between them, and then NMS has been used to eliminate the ones with the lowest confidence scores. For each overlapping character, more than one character will have some overlapping portions. OV is used to determine to how much extent the characters overlap. The smaller the value of OV for a character, the more overlap there is with another character.
{\fontsize{9.5}{11}
\begin{equation}
\text{OV} = \frac{\text{Character's Personal Area}}{\text{Common Area of All Overlapping Characters}} \label{eq1}
\end{equation}
}
The authors empirically decided that if the value of OV for any overlapping character is less than 2, more than one class strongly share a common area. The class with the highest confidence score will be selected in this case. In this way, the models detect a single class for one character/diacritic. 
In Fig. \ref{fig3c}, the bounding boxes of \textbf{\bengalifont 'ষ,' 'ট,' 'ষ্ট'} have an area of 0.04, 0.07, and 0.14 respectively.\textbf{\bengalifont 'ষ্ট'} has a common area of 0.03 with \textbf{\bengalifont 'ষ'} and 0.065 with \textbf{\bengalifont 'ট'}. The OV for  \textbf{\bengalifont 'ষ'}, and \textbf{\bengalifont 'ট'} is respectively  1.33 and 1.08. As both of their value is less than 2, it means both have lots of common/overlapping portions with  \textbf{\bengalifont 'ষ্ট'}. It concludes that NMS needs to eliminate two out of three as all three are different results for one character. \textbf{\bengalifont 'ষ,' 'ট,'} and \textbf{\bengalifont 'ষ্ট'} have confidence scores of 43\%, 35\%, and 89\% respectively. So, \textbf{\bengalifont 'ষ্ট'} has been chosen from these three as it has the highest confidence score. (Fig. \ref{fig3d})

\subsubsection{Preserving Necessary Overlap(PNO)}
\begin{figure}
  \centering
  \resizebox{0.5\textwidth}{!}{
      \begin{subfigure}[b]{0.45\textwidth}
        \centering
        \includegraphics[width=\textwidth,height=80pt]{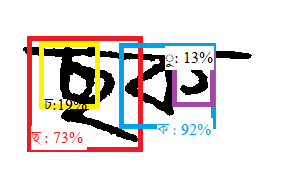}
        \captionsetup{font=LARGE}
        \caption{Result Classes With All Scores}
        \label{fig3a}
      \end{subfigure}
      \begin{subfigure}[b]{0.45\textwidth}
        \centering
        \includegraphics[width=\textwidth,height=80pt]{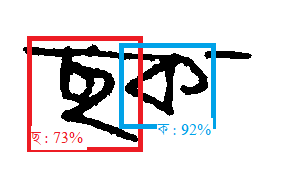}
        \captionsetup{font=LARGE}
        \caption{Eliminating Low Scores}
        \label{fig3b}
      \end{subfigure}
      \begin{subfigure}[b]{0.45\textwidth}
        \centering
        \includegraphics[width=\textwidth,height=80pt]{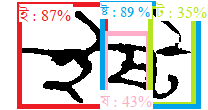}
        \captionsetup{font=LARGE}
        \caption{Before NMS}
        \label{fig3c}
      \end{subfigure}}
    \resizebox{0.5\textwidth}{!}{
      \begin{subfigure}[b]{0.45\textwidth}
        \centering
        \includegraphics[width=\textwidth,height=80pt]{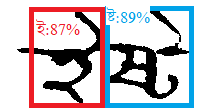}
        \captionsetup{font=LARGE}
        \caption{After NMS}
        \label{fig3d}
      \end{subfigure}
      \begin{subfigure}[b]{0.45\textwidth}
        \centering
        \includegraphics[width=\textwidth,height=80pt]{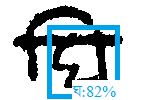}
        \captionsetup{font=LARGE}
        \caption{Before PNO}
        \label{fig3e}
      \end{subfigure}
      \begin{subfigure}[b]{0.45\textwidth}
        \centering
        \includegraphics[width=\textwidth,height=80pt]{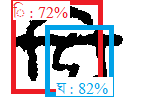}
        \captionsetup{font=LARGE}
        \caption{Result After PNO}
        \label{fig3f}
      \end{subfigure}
  }
  \caption{Results of SSD before and after eliminating classes with lower confidence scores, applying NMS and PNO}
  \label{fig4}
\end{figure}

There are some cases where NMS is troublesome rather than a blessing. E.g., in Bangla, there are some diacritics such as \textbf{\bengalifont 'ি,' ' ী,' ' ৌ'}, for which overlapping is obvious. In these cases, if NMS is applied as usual, either a consonant or the diacritic will be eliminated. That is why, whenever these diacritics are involved along with the blue labeled consonants/conjuncts from TABLE \ref{table1} in any overlapping issue, the threshold value for OV is decreased to 0.5 instead of 2(empirically decided).In Fig.\ref{fig3e}, the bounding boxes of \textbf{\bengalifont 'ি'} and \textbf{\bengalifont 'ঘ'} have an area of 0.20 and 0.13 and their common area is 0.09. \textbf{\bengalifont 'ি'} and \textbf{\bengalifont 'ঘ'} have OV of 2.22 and 1.44. After NMS, \textbf{\bengalifont 'ি'} is eliminated as they share an OV value less than 2. However, decreasing the OV threshold to 0.5 helps preserve both \textbf{\bengalifont 'ি'} and \textbf{\bengalifont 'ঘ'}. (Fig. \ref{fig3f})
Also, another similar case is with \textbf{\bengalifont ' ঁ'} and characters which have some extended part over matra like \textbf{\bengalifont 'ট,' 'ঠ,' and 'ষ্ট'}. In this case, the OV threshold is also 0.5.
Bangla diacritics like \textbf{\bengalifont 'ে,' 'ৈ'} could also be added in this step, but in different handwriting style, these diacritics look very much like the Bangla consonants  {\bengalifont 'ট,' 'ঢ,' 'চ'}. This makes the models detect at least two classes for one character. That is why these diacritics are not added in this step despite creating significant overlaps in some cases.

\subsubsection{Text Words}
The text word is produced by processing and rearranging detected characters according to their outcome. Characters are rearranged based on their coordinate value. Some characters and diacritics such as \textbf{\bengalifont' ে', 'ি,' 'ৈ,' 'ৌ,' '্র'} are written before their corresponding character but typed after it. \textbf{\bengalifont 'আ'}, once again, was not one of the identified classes. When the characters \textbf{\bengalifont 'অ'} and \textbf{\bengalifont 'া'} are found together, they are substituted with \textbf{\bengalifont 'আ’}\cite{id18}. This type of exception is taken into account while reordering the characters.

\section{Results}
\subsection{Datasets}
The authors created a customized dataset of 9841 handwritten Bangla word images with one or multiple characters for the Word Recognition stage. 70\% of the images of the dataset are collected from different people's handwriting, and 30\% of the word images are collected from the CMATERdb 3.1.1\cite{id21}, CMATERdb 2.1.2.1, CMATERdb 3.1.3.1\cite{id22}, CMATERdb 3.1.3.3\cite{id23}.  The inclusion of various CMATERdb datasets is to make the customized dataset more versatile. After collecting the images of papers containing many words from different writers, the words have been segmented and binarized\cite{id24}so that each image includes only a word with one or more characters/classes. Even after segmenting the words, the word images have been manually fine-tuned to eliminate irrelevant information besides word texts. Then the characters from the word images are labeled manually according to their corresponding classes.

Out of 9841-word images from the customized dataset, 7701 images are used for training and 2140 for testing. In the 7701 train images, there is a total of 18526 character/diacritic/number/punctuation instances. The words for train images are selected in such a way that the frequent Bangla characters used daily have the most occurrences. In the 2140 test images, there are 5451 character/diacritic/number/punctuation instances. The words for test images are selected in such a way that every class has at least 50 instances so that the result does not create any biases.

\subsection{Evaluation Metrics}
Various metrics such as Recall\eqref{eq1}, Precision\eqref{eq2}, and F-measure (F1-score)\eqref{eq3} are used to measure the system's recognition rate accuracy on character level. Macro-averaging has been employed to summarize the models' performance across different character classes. 
{\fontsize{9.5}{9}
\begin{equation}
\text{Recall} = \frac{\text{TP}}{\text{TP + FN}} \label{eq1}
\end{equation}
\begin{equation}
\text{Precision} = \frac{\text{TP}}{\text{TP + FP}} \label{eq2}
\end{equation}
\begin{equation}
\text{F1-score} = 2 \times \frac{\text{Precision} \times \text{Recall}}{\text{Precision} + \text{Recall}} \label{eq3}
\end{equation}
}
Here, True Positive(TP) refers to a character class that is correctly identified and recognized. A false negative(FN) indicates that a class/character that is present was not detected; something was overlooked. False Positive(FP) occurs when a character in an image is correctly identified but incorrectly classified.

\subsection{Experimental Results on Character Level}
In this experiment, the test images are tested separately by the three different models, and Recall, Precision, and F1-score are measured and reported in TABLE \ref{table2}.
% \begin{table}[b]
%   \centering
%   \caption{Recall, Precision, and F1-Score of Proposed Work Applied on Different Models}
%   \renewcommand{\arraystretch}{1.5}
%   \begin{tabularx}{0.5\textwidth}{|m{2.05cm}|m{1.75cm}|m{1.75cm}|m{1.75cm}|}
%     \hline
%     \centering Model & \centering Recall (\%) & \centering Precision(\%) & \centering F1 score (\%) \tabularnewline
%     \hline
%     \centering SSD with MobileNetV2 & \centering $95.88$ & \centering $87.62$ & \centering $91.19$ \tabularnewline
%     \hline
%     \centering Faster R-CNN with InceptionResNetV2 & \centering $92.52$ & \centering $86.51$ & \centering $88.25$ \tabularnewline
%     \hline
%     \centering Ensemble model & \centering $96.83$ & \centering $89.42$ & \centering $92.61$ \tabularnewline
%     \hline
%   \end{tabularx}
  
%   \label{table2}
% \end{table}
\begin{table}[b]
  \centering
  \caption{Comparative Evaluation: Recall, Precision, and F1-Score of the proposed method on character level and comparison with existing approaches}
  \renewcommand{\arraystretch}{1.5}
  \begin{tabularx}{0.5\textwidth}{|m{1.30cm}|m{1.9cm}|m{1.1cm}|m{1.1cm}|m{1.45cm}|}
    \hline
    \centering Approach & \centering Model & \centering Recall (\%) & \centering Precision (\%) & \centering F1 score (\%) \tabularnewline
    \hline
    \multirow{3}{*}[-4ex]{\textbf{Proposed}} & SSD with MobileNetV2 & \centering $95.88$ & \centering $87.62$ & \centering $91.19$ \tabularnewline
    \cline{2-5}
    &\centering Faster R-CNN with InceptionResNetV2 & \centering $92.52$ & \centering $86.51$ & \centering $88.25$ \tabularnewline
    \cline{2-5}
    &\centering \textbf{Ensemble model} & \centering \textbf{96.83} & \centering \textbf{89.42} & \centering \textbf{92.61} \tabularnewline
    \hline
    \multirow{2}{*}[-3ex]{Satter\cite{id24}} & SSD with MobileNetV2 & \centering $88.02$ & \centering $68.3$ & \centering $74.04$ \tabularnewline
    \cline{2-5}
    &\centering Faster R-CNN with InceptionResNetV2 & \centering $92.18$ & \centering $80.20$ & \centering $83.72$ \tabularnewline
    \hline
    \centering Majid et al.\cite{id17} &\centering Faster R-CNN with VGG-16 & \centering $91.24$ & \centering $88.25$ & \centering $89.72$ \tabularnewline
    \hline
    \centering Hasan et al.\cite{id18} &\centering SSD with MobileNetV2 & \centering $80.43$ & \centering $88.26$ & \centering $84.21$ \tabularnewline
    \hline
  \end{tabularx}
  
  \label{table2}
\end{table}

Here, both of the models have their limitations. Overall, SSD gives better results than Faster R-CNN, but there are cases where SSD fails frequently, e.g., SSD classifies \textbf{\bengalifont 'ঐ'} every time as \textbf{\bengalifont 'এ'}. Also, SSD often fails to classify the numbers correctly. Faster R-CNN does not have difficulty classifying these characters but gives poor results where diacritics are involved. However, the ensemble model takes both of their results and gives the best result with the highest confidence score, and that is why the ensemble model has an improved result over these two. The ensemble model having a recall of 96.83\% means only 3.17\% of the actual classes were not identified. It shows that the proposed model significantly overcame the challenges of the cursive nature of Bangla handwriting.

Satter\cite{id24} introduced a similar approach using both models where no ensemble method was used, and SSD and Faster R-CNN had an F1-score of 74.04\% and 83.72\%. Majid et al.\cite{id17} had a similar work using two Faster R-CNNs, and they found a recall and precision of 91.24\% and 88.25\% after merging both the models' result to form a transcription. However, it's worth noting that the F1-score they reported was formulated using the geometric mean of precision and recall. If they had calculated the F1-score using (4) (harmonic mean of precision and recall), it would have been 89.72\%. Hasan et al.\cite{id18} worked on a similar approach using SSD; their F1-score for character recognition was 84.21\%, and they did not even include any conjunct character. All of their results have been shown in TABLE \ref{table2} to display a comparison between the proposed methodology and some of the existing methodologies. 

\subsection{Experimental Results on Word Level}
Levenshtein Distance\cite{id24} is used to measure the models' performance on word level on 2140 test words. 100\% accuracy means the predicted and annotated words are precisely the same. 75\% accuracy means if there are four annotated classes in the actual output, the model correctly predicts 3 of the classes. TABLE \ref{table3} shows the models' cumulative accuracy using Levenshtein Distance. TABLE \ref{table4} shows the effectiveness of the NMS and PNO steps of the proposed methodology by showing the result if they were omitted.
\begin{table}[b]
  \centering
  \caption{Accuracy on word level for different models}
  \renewcommand{\arraystretch}{1.5}
  \begin{tabularx}{0.5\textwidth}{|m{1.28cm}|m{1.5cm}|m{2.1cm}|m{2.42cm}|}
    \hline
    \centering Minimum Accuracy & \centering \% of Words (SSD) & \centering {\footnotesize\% of Words (Faster R-CNN)} & \centering \textbf{{\footnotesize\% of Words (Ensemble Model)}} \tabularnewline
    \hline
    \centering $100$ & \centering $70.89$ & \centering $64.07$ & \centering \textbf{72.52} \tabularnewline
    \hline
    \centering $>=75$ & \centering $77.57$ & \centering $70.93$ & \centering \textbf{78.56} \tabularnewline
    \hline
    \centering $>=50$ & \centering $92.29$ & \centering $88.83$ & \centering \textbf{93.04} \tabularnewline
    \hline
    \centering $>=25$ & \centering $95.84$ & \centering $93.60$ & \centering \textbf{96.12} \tabularnewline
    \hline
    \centering $>=0$ & \centering $100$ & \centering $100$ & \centering \textbf{100} \tabularnewline
    \hline
    
  \end{tabularx}
  
  \label{table3}
\end{table}

\begin{table}[ht]
  \centering
  \caption{Accuracy on word level for different models if NMS and PNO were omitted}
  \renewcommand{\arraystretch}{1.5}
  \begin{tabularx}{0.5\textwidth}{|m{1.28cm}|m{1.2cm}|m{1.6cm}|m{1.2cm}|m{1.6cm}|}
    \hline
    \centering Scenario &  \multicolumn{2}{|m{2.8cm}|}{If only PNO was omitted} & \multicolumn{2}{m{2.8cm}|}{If both NMS and PNO were omitted}   \tabularnewline
    \hline
    \centering Minimum Accuracy & \centering \% of Words (SSD) & \centering \% of Words (Faster R-CNN) & \centering \% of Words (SSD) & \centering \% of Words (Faster R-CNN)  \tabularnewline
    \hline
    \centering $100$ & \centering $68.64$ & \centering $62.43$  & \centering $30.79$ & \centering $13.97$ \tabularnewline
    \hline
    \centering $>=75$ & \centering $76.36$ & \centering $70.28$  & \centering $33.13$ & \centering $16.45$\tabularnewline
    \hline
    \centering $>=50$ & \centering $92.38$ & \centering $88.79$ & \centering $50.23$ & \centering $28.64$ \tabularnewline
    \hline
    \centering $>=25$ & \centering $96.12$ & \centering $93.97$ & \centering $58.64$ & \centering $36.45$ \tabularnewline
    \hline
    \centering $>=0$ & \centering $100$ & \centering $100$ & \centering $100$ & \centering $100$ \tabularnewline
    \hline
    
  \end{tabularx}
  
  \label{table4}
\end{table}

Some word images in the test dataset contain only a single character. So, if a model fails to detect/classify that character, the word recognition accuracy directly falls into the 0\% accuracy category. That is why there are many words with 0\% recognition accuracy. Also, there are many small words (e.g., \textbf{\bengalifont 'এই,' 'ছন্দ,' 'যে' }). If a model wrongly classifies any character, the word is 50\% accurate. These types of instances create misinterpretations about the models' performance on the word level. This misinterpretation is visible if the character recognition result is compared with this result.

Hossain et al.\cite{id31} found a word recognition accuracy of 75\% when handwriting is poor. Also, they tested against only 100-word images and did not include any conjuncts and their system highly relied on the alignment of the Matra Row. On the other hand, the proposed method does not have any reliance on the Matra Row's alignment and character sizes, and the ensemble model can recognize 96.12\% of the test words to some extent.

\section{Conclusion}
OCR systems use computer vision to recognize and transform word images into editable text. In this research, a word recognition system for open vocabulary handwritten Bangla words is created, which uses a bespoke dataset and deep learning models to recognize words from images. However, the system has limitations too, e.g., only 17 conjunct characters have been included here, but almost 200 more conjunct characters can be included. Besides, Bangla has more punctuation marks than \textbf{\bengalifont '।'}. Also, a post-processing technique can be introduced to correct the errors generated in the texts.

The object detection models in the proposed method have been trained with a wide variety of handwriting. Also, it is not limited to any closed vocabulary. So it can recognize different people's writing. The most promising fact about the proposed method is that the ensemble method takes the result of two parallelly trained object detection models and gives the best result. The improved result of the ensemble model shows a new pathway for other researchers. The dataset used here can help future researchers as creating and labeling a huge dataset takes a lot of time. Also, the introduction of NMS and PNO helps improve the result.

The system's three outputs have been compared, and the performance has been measured on character and word level. The recognized words can be further used to develop a post-processing word correction for them. The existing system may help transition to a more established OCR system.

% \section{Acknowledgment} 
% We extend our heartfelt gratitude to all volunteers who contributed to providing diverse handwriting and label images for the Dataset project. Special recognition goes to Sajia Abedin, Mishu Baidya, Md Kaisarul Islam, MD.Abu Loman Hossain Shuvo. We sincerely thank the Center for Microprocessor Application for Training, Education, and Research (CMATER) group for their support.

\bibliographystyle{IEEEtran}
\bibliography{references}

\end{document}